\documentclass[sigconf]{acmart}
\renewcommand\footnotetextcopyrightpermission[1]{} % removes footnote with conference information in first column
\AtBeginDocument{%
  \providecommand\BibTeX{{%
    \normalfont B\kern-0.5em{\scshape i\kern-0.25em b}\kern-0.8em\TeX}}}

\setcopyright{none}
\usepackage{todonotes}
\usepackage{multirow}

\begin{document}
\pagestyle{plain} % removes running headers
%%
%% The "title" command has an optional parameter,
%% allowing the author to define a "short title" to be used in page headers.
\title{MultiFIX: An XAI-friendly feature inducing approach to building models from multimodal data}
% Learning Exlainable Features Simultaneously from Multimodal Data
% Making Multimodal Data Fusion Explainable and Transparent

%%
%% The "author" command and its associated commands are used to define
%% the authors and their affiliations.
%% Of note is the shared affiliation of the first two authors, and the
%% "authornote" and "authornotemark" commands
%% used to denote shared contribution to the research.
\author{Mafalda Malafaia}
\email{Mafalda.Malafaia@cwi.nl}
\orcid{0002-8081-0454}
\affiliation{%
  \institution{Centrum Wiskunde \& Informatica}
  \city{Amsterdam}
  \country{The Netherlands}
}

\author{Thalea Schlender}
\email{T.Schlender@lumc.nl}
\affiliation{%
  \institution{Leiden University Medical Center}
  \city{Leiden}
  \country{The Netherlands}}

\author{Peter A.N. Bosman}
\email{Peter.Bosman@cwi.nl}
\affiliation{%
  \institution{Centrum Wiskunde \& Informatica}
  \city{Amsterdam, The Netherlands}\\
  \institution{Delft University of Technology}
  \city{Delft}
  \country{The Netherlands}
}

\author{Tanja Alderliesten}
\email{T.Alderliesten@lumc.nl}
\affiliation{%
  \institution{Leiden University Medical Center}
  \city{Leiden}
  \country{The Netherlands}}

%%
%% By default, the full list of authors will be used in the page
%% headers. Often, this list is too long, and will overlap
%% other information printed in the page headers. This command allows
%% the author to define a more concise list
%% of authors' names for this purpose.
\renewcommand{\shortauthors}{Malafaia, et al.}

%%
%% The abstract is a short summary of the work to be presented in the
%% article.
\begin{abstract}

In the health domain, decisions are often based on different data modalities. Thus, when creating prediction models, multimodal fusion approaches that can extract and combine relevant features from different data modalities, can be highly beneficial. Furthermore, it is important to understand how each modality impacts the final prediction, especially in high-stake domains, so that these models can be used in a trustworthy and responsible manner. We propose MultiFIX: a new interpretability-focused multimodal data fusion pipeline that explicitly induces separate features from different data types that can subsequently be combined to make a final prediction. An end-to-end deep learning architecture is used to train a predictive model and extract representative features of each modality. Each part of the model is then explained using explainable artificial intelligence techniques. Attention maps are used to highlight important regions in image inputs. Inherently interpretable symbolic expressions, learned with GP-GOMEA, are used to describe the contribution of tabular inputs. The fusion of the extracted features to predict the target label is also replaced by a symbolic expression, learned with GP-GOMEA. Results on synthetic problems demonstrate the strengths and limitations of MultiFIX. Lastly, we apply MultiFIX to a publicly available dataset for the detection of malignant skin lesions.

%In the health sector, several decision-making processes rely on contributions from different data modalities. Thus, multimodal fusion approaches can be highly beneficial in extracting and combining relevant features for a final prediction. Furthermore, it is important to understand how each modality impacts the final decision, especially in high-stake domains, so that these models can be used in a trustworthy and responsible manner. We propose MulTi-FIX: a new interpretability-focused multimodal data fusion pipeline to simultaneously learn features from several data modalities for a final prediction. An end-to-end deep learning architecture is used to train the predictive models and extract representative features of each modality. These are then explained using explainable artificial intelligence techniques: attention maps that highlight important regions in image inputs, and inherently interpretable symbolic expressions that are learned with GP-GOMEA to translate the contribution of tabular inputs. The fusion of the extracted features to obtain the target label is replaced by a symbolic expression with GP-GOMEA. As a proof-of-concept for this innovative pipeline, synthetic problems demonstrate the strengths and limitations to solve representative problems. Lastly, we showcase the application of the pipeline to a publicly available medical dataset for the detection of malignant skin lesions.
%Lastly, to showcase the potential of this pipeline, a publicly available medical dataset is used to detect melanoma and analyse the importance of the inputs.

\end{abstract}

%%
%% The code below is generated by the tool at http://dl.acm.org/ccs.cfm.
%% Please copy and paste the code instead of the example below.
%%

%\ccsdesc[500]{Do Not Use This Code~Generate the Correct Terms for Your Paper}
%\ccsdesc[300]{Do Not Use This Code~Generate the Correct Terms for Your Paper}
%\ccsdesc{Do Not Use This Code~Generate the Correct Terms for Your Paper}
%\ccsdesc[100]{Do Not Use This Code~Generate the Correct Terms for Your Paper}

%%
%% Keywords. The author(s) should pick words that accurately describe
%% the work being presented. Separate the keywords with commas.

%\received{20 February 2007}
%\received[revised]{12 March 2009}
%\received[accepted]{5 June 2009}

%%
%% This command processes the author and affiliation and title
%% information and builds the first part of the formatted document.
\maketitle

\section{Introduction}

Within Machine Learning~(ML), the focus of multimodal fusion approaches is on the integration of various forms of data. Empirical evidence suggests that multimodal ML may outperform unimodal approaches~\cite{huang2022review}, and is deemed more robust because it is capable of leveraging a larger volume of information~\cite{nature_multimodal_review22}. From a practical standpoint, it is also highly desirable to leverage all available data to achieve the most accurate predictions, e.g., in medicine one would want to consider medical images, patient characteristics, and lab measurements all at once, just like a physician would.

Most state-of-the-art multimodal approaches make use of Deep Neural Networks~(DNNs), which have been proven successful in several different tasks~\cite{azam2022review}. However, the opaque and complex nature of DNNs limits their usability in critical domains, such as the medical sector. Multiple factors, including legal and privacy aspects, among others, strongly encourage that artificial intelligence frameworks have to be human-verifiable and interpretable to be deployed
in real-life situations~\cite{justify_xai_17}. 

The field of evolutionary computation can play an important role here as Evolutionary Algorithms (EAs) can be used to optimise complex, discrete and continuous, problems, offering a wider range of possibilities to train compact but explainable models that cannot be trained (well) through gradient-based methods. A key example thereof is Genetic Programming~(GP) which can be used to learn, for instance, small expressions as part of feature engineering~\cite{tran2017using, virgolin2020explaining}. However, GP is less well suited to perform image analysis tasks than modern Deep Learning~(DL). Such models, while opaque, still permit various forms of post-hoc explainability, marking what parts of an image were crucial for certain predictions. Ideally, one would want to leverage the strengths of explainable DL models for images and inherently explainable (GP) models for tabular data in a multimodal fashion. However, this is non-trivial since the training procedures of said models are typically different, hampering joint minimisation of a loss function. Moreover, ideally, one would want to be able to identify how features that are themselves explainable or inherently interpretable and are constructed from individual modalities contribute, potentially together with features from other modalities to a final prediction.

Current interpretable approaches for multimodal learning still cannot achieve all of the above. Notable mentions are~\citet{wang2021interpretability} for skin lesion diagnosis and~\citet{azher2023assessment} for cancer prognostic detection, which focus only on using \textit{post-hoc} methods for explainability.
 
Thus, we propose MultiFIX: Multimodal Feature Inducing eXplainable artificial intelligence to build interpretable models that use several data modalities such that the contribution of each modality to the final prediction can be identified clearly, maximising explainability potential. More specifically, with MultiFIX we aim to unite the following advantages from the interpretability perspective: 

\begin{itemize}
    \item \textbf{Embedded feature engineering} to induce representative feature(s) per modality before fusion. 
    \item \textbf{Explainability of engineered features} consisting of contribution heatmaps for image-induced features and interpretable expressions for tabular-induced features.
    \item \textbf{Explainability of data modality fusion} consisting of interpretable expressions for the final feature-combining prediction block to perform target prediction.
\end{itemize}

\vspace*{-2mm}
\section{MultiFIX}
\label{sec:framework}

\begin{figure*}
    \centering
    \scalebox{0.85}{\includegraphics{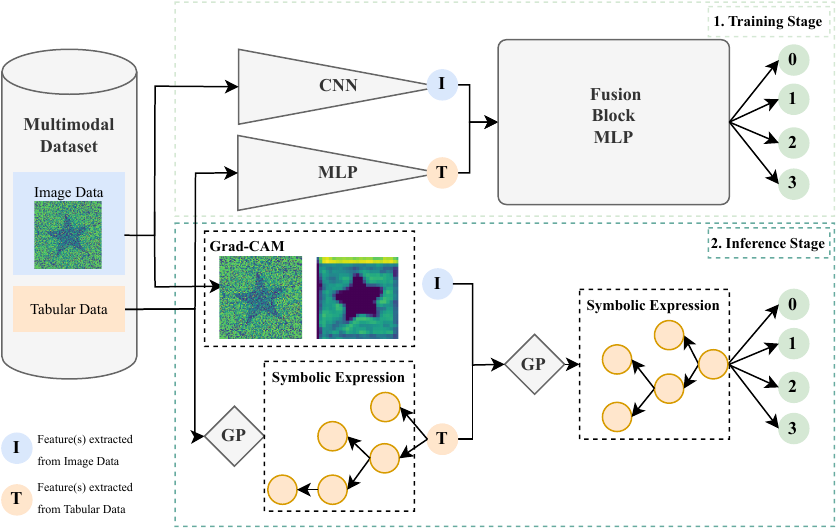}}
    \vspace*{-3mm}
    \caption{Overview of MultiFIX. The available data is given as input to the feature-inducing blocks (NNs in this paper). The output thereof is passed into a fusion block to obtain the final prediction. Representative features (I from image data and T from tabular data) - are thereby learned simultaneously when training the entire architecture (top). After the Training Stage, in the Inference Stage, induced image features are explained through Grad-CAM, and symbolic expressions are obtained for both the tabular features and the fusion block with GP-GOMEA. The GP-GOMEA models can also be used to replace their NN counterparts, making the models more than explanations but rather an integral part of the final model, increasing its verifiability potential. In the present figure, the Multiclass Problem is used to illustrate MultiFIX.}
    \label{fig:framework}
\end{figure*}

With MultiFIX, we aim to simultaneously learn key features from the given data modalities to make predictions. MultiFIX consists of one feature-inducing block for each data type and one fusion block that combines the induced features to make final predictions. Different types of (ML) models can be used in each block, but here we showcase MultiFIX in combination with DL models for each block at first, with the option to replace them with other, explainable, models afterward, or using post-hoc explainability techniques for a DL model. This is particularly convenient for training, as then end-to-end learning can be employed using standard DNN techniques. An overview of MultiFIX is provided in Figure~\ref{fig:framework}.

Gradient-weighted Class Activation Mapping~(Grad-CAM)~\cite{grad_cam19} is a technique that uses the gradient information in the convolutional layers of a DL model and generates visual explanations that are correlated with the image contribution for a specific target. GP-GOMEA~\cite{gp_gomea21} is a model-based evolutionary algorithm for GP known for its efficiency in evolving small and, thus, potentially interpretable symbolic expressions~\cite{srbench}. The feature engineering blocks are interpreted according to the nature of the data: Grad-CAM is used to obtain explanatory heatmaps for the image block; GP-GOMEA is used to obtain a symbolic expression that represents the tabular feature(s). The GP model then replaces the NN model. Finally, GP-GOMEA is used once again to find a small symbolic expression to replace the fusion block.

We have adapted GP-GOMEA to handle a mixture of numeric and Boolean operators, as well as the possibility of using if-then-else statements to model discontinuities\textcolor{green}{~\cite{thalea_gpg}}. The GP-GOMEA settings used throughout the present work are described in Table~\ref{tab:gpg_settings}. Each GP-GOMEA experiment is run for 5 seeds. The symbolic expression pertaining to the run that achieved the best accuracy is chosen.

In the remainder of the paper, MultiFIX is tested on various synthetic problems and a medical use case. In each case, 5-fold cross-validation is utilised to ensure the generalisation capability of the models. Single-modality performance is used as a baseline to compare the improvements when combining data modalities. 

\begin{table}[]
\begin{tabular}{ll}
\textbf{Population size} & initially 64 (using IMS~\cite{gp_gomea21}) \\ \hline
\textbf{Number of generations } & 512 \\ \hline
\textbf{Operators} & numeric and Boolean; if-then-else \\ \hline
\textbf{Maximum tree depth} & {[}2, 3, 4{]}
\end{tabular}
\caption{GP-GOMEA settings used.}
\label{tab:gpg_settings}
\vspace*{-6mm}
\end{table}

\section{Proof-of-concept experiments}
\label{sec:dummy}

To study and evaluate the potential of MultiFIX, several synthetic problems were created to represent different correlations between input modalities and end-points. For all problems, representative features are present in each modality. MultiFIX is in principle trained end-to-end to predict the final classification that needs both inputs. The underlying features of each input are not provided to MultiFIX, and therefore, are trained in an unsupervised fashion. According to the complexity of each problem, alternative approaches are used to train MultiFIX and address current shortcomings.

\subsection{Multiclass Problem}
\label{subsec:multiclass}
In this problem, one representative feature from each modality is used to predict an end-point that varies among 4 possible values according to the combination of the features in the two modalities. While each modality is slightly correlated with the end-point, complementary information of both modalities is needed for optimal predictions. For the input-output table of this problem, see Table~\ref{tab:multiclass_table}.

\begin{table}
    \centering
    \begin{tabular}{c c|c}
         $ft_{square}$ & $ft_{A}$ & $Y_{multiclass}$\\ \hline
         0 & 0 & 0 \\
         0 & 1 & 1 \\
         1 & 0 & 2 \\
         1 & 1 & 3
    \end{tabular}
%    \caption{Definition of the end-point prediction in the Multiclass Problem. The target cannot be predicted with total certainty based on solely the images,  $ft_{square}$, or solely  the feature extracted from the tabular data, $ft_{A}$.}
    \caption{Definition the Multiclass Problem based on image feature $ft_{square}$ and tabular feature $ft_{A}$.}
    \label{tab:multiclass_table}
\vspace*{-6mm}
\end{table}

\subsubsection{Data Description}
\begin{figure*}
    \centering
    \scalebox{0.8}{\includegraphics[width=0.7\textwidth]{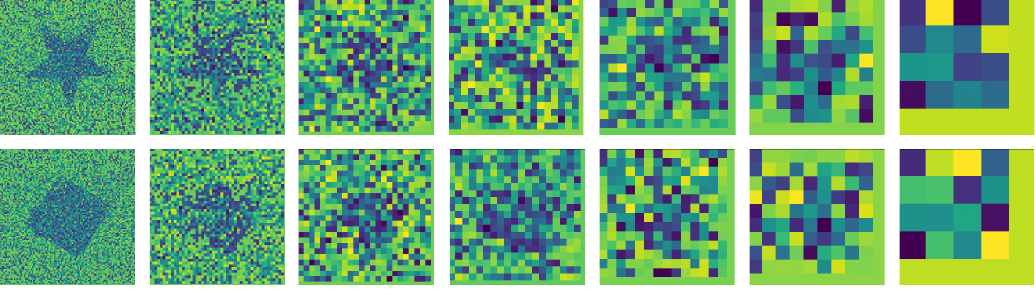}}
\vspace*{-3mm}
\caption{Representative samples of input images for the Multiclass Problem. The first row displays images belonging to the star class (label $0$), and the second row displays images from the square class (label $1$). Each column contains images with a different resolution, starting with the image with $100\times100$ pixels, following $50\times50$, $25\times25$, $20\times20$, $15\times15$, $10\times10$ and, lastly, $5\times5$ pixels.}
    \label{fig:multiclass_samples}
\vspace*{-3mm}
\end{figure*}

For the image modality, a publicly available dataset from \textit{Kaggle}~\cite{star_dataset} was adapted to obtain images with two possible shapes: a star (label $0$) or a square (label $1$), as illustrated in Figure~\ref{fig:multiclass_samples}. Because real-world images are not without noise, random Gaussian noise of different magnitudes was added to the images. For the tabular modality, a \textit{scikit-learn} package~\cite{scikit-learn} was utilized to generate a random binary classification problem with 20 continuous features of which 10 are informative (i.e., relevant for the prediction), 5 are linear combinations of the informative features and, thus, redundant, and 5 are random (i.e., useless for the prediction). These features were generated along with a binary label A, which constitutes the feature that we want to extract from the tabular modality and that is ultimately important for the final prediction. 200 samples were created for each data modality.

To study the influence of the input quality in MultiFIX, both data types were modified to decrease their ability to represent the intermediate feature. Images were resampled into lower resolutions: $100\times100$, $50\times50$, $25\times25$, $20\times20$, $15\times15$, $10\times10$, and $5\times5$ pixels. Representative image samples are available in Figure~\ref{fig:multiclass_samples}. Gaussian noise was added to the tabular samples with using different standard deviations: $0$, $2.5$, $5$, $7.5$, $10$, $15$, and $20$.

\subsubsection{Experimental Design}

A pre-trained Resnet~\cite{he2016deep} is used for the image data and a Multilayer Perceptron~(MLP) is used for the tabular data as well as the fusion block. For the tabular data, the MLP has two hidden layers. For the fusion block, the number of hidden layers is a hyperparameter. These NNs are also used for single-modality models.

MultiFIX is implemented following the description in Section~\ref{sec:framework}. Additionally, the training stage includes fine-tuning in two steps: Hyper-Parameter Optimisation~(HPO), in which the Learning Rate (LR) and the Weight Decay~(WD) are optimised using grid-search; Neural Architectural Search~(NAS), using an open-source library~\cite{nni2021} that leverages the Tree-Structured Parzen Estimator~\cite{tpe2013}, in which the portion of the architecture dedicated to the fusion of the features is fine-tuned with a small model search space that includes the number of hidden layers, number of nodes per layer, batch normalisation, and dropout. The search space is described in Table~\ref{tab:multiclass_grid}. The Adam optimiser and Cross Entropy Loss are used for training.

\begin{table}[]
\scalebox{0.81}{
\begin{tabular}{l|ll}
\multirow{2}{*}{\textbf{HPO}} & \textbf{LR} & {[}1e-2, 5e-3, 1e-3, 5e-4{]} \\ \cline{2-3} 
 & \textbf{WD} & {[}1e-2, 1e-3, 0{]} \\ \hline
\multirow{4}{*}{\textbf{NAS}} & \textbf{ft block activation} & {[}ReLU, Sigmoid{]} \\ \cline{2-3} 
 & \textbf{hidden layers} & {[}1, 2{]} \\ \cline{2-3} 
 & \textbf{dropout (per layer)} & {[}0, 0.125, 0.25{]} \\ \cline{2-3} 
 & \textbf{width (per layer)} & {[}16, 32, 64{]}
\end{tabular}}
\caption{Optimisation grid for Multiclass Problem. The best set of parameters is chosen according to the loss (lowest average $\pm$ standard deviation calculated over the 5 folds).}
\label{tab:multiclass_grid}
\vspace*{-8mm}
\end{table}

At the end of the training stage, training is done once more with the best hyperparameters to validate the results, save the best model, and apply interpretability methods. Grad-CAM is used on the activations from the first residual convolutional block of the Resnet. GP-GOMEA is used with the settings in Table~\ref{tab:gpg_settings}.

\subsubsection{Performance Results}
First, we look at the potential of how easily the required intermediate features can be learned by each feature engineering block. All results are measured using Balanced Accuracy~(BAcc) and are presented in Tables~\ref{tab:img_ft_multiclass}~and~\ref{tab:tab_ft_multiclass}. While it shows that the features can be represented by the blocks, it is important to keep in mind that within MultiFIX, these features are expected to be learned in an unsupervised fashion. 

\begin{table*}
\scalebox{0.95}{
\tabcolsep=1mm
\begin{tabular}{c|ccccccc}
    \textbf{\begin{tabular}[c]{@{}c@{}}Image Resolution ($pixels\times pixels$)\end{tabular}} & $100\times100$ & $50\times50$ & $25\times25$ & $20\times20$ & $15\times15$ & $10\times10$ & $5\times5$\\ \hline
    %\textbf{AUC} & 1.000 $\pm$ 0.000 & 1.000 $\pm$ 0.000 & 0.947 $\pm$ 0.045 & 0.914 $\pm$ 0.024 & 0.840 $\pm$ 0.043 & 0.608 $\pm$ 0.078 & 0.557 $\pm$ 0.036 \\
    \textbf{BAcc}  & 1.000 $\pm$ 0.000 & 1.000 $\pm$ 0.000 & 0.825 $\pm$ 0.052 & 0.835 $\pm$ 0.041 & 0.735 $\pm$ 0.041 & 0.515 $\pm$ 0.044 & 0.560 $\pm$ 0.034
\end{tabular}
}
\caption{Performance results (average $\pm$ std.dev. over 5 folds) for supervised image feature extraction in the Multiclass Problem.}
\label{tab:img_ft_multiclass}
\vspace*{-6mm}
\end{table*}

\begin{table*}[]
\scalebox{0.81}{
\begin{tabular}{c|ccccccc}
\textbf{\begin{tabular}[c]{@{}c@{}}Gaussian Noise Std.Dev.\end{tabular}} & 0 & 2.5 & 5 & 7.5 & 10 & 15 & 20 \\ \hline
%\textbf{AUC} & 1.000 $\pm$ 0.000 & 0.920 $\pm$ 0.036 & 0.771 $\pm$ 0.037 & 0.610 $\pm$ 0.058 & 0.544 $\pm$ 0.060 & 0.470 $\pm$ 0.067 & 0.521 $\pm$ 0.067 \\
\textbf{BAcc} & 0.980 $\pm$ 0.024 & 0.865 $\pm$ 0.020 & 0.700 $\pm$ 0.050 & 0.605 $\pm$ 0.064 & 0.475 $\pm$ 0.050 & 0.480 $\pm$ 0.029 & 0.515 $\pm$ 0.086
\end{tabular}
}
\caption{Performance results (average $\pm$ std.dev. over 5 folds) for supervised tabular feature extraction in the Multiclass Problem.}
\label{tab:tab_ft_multiclass}
\vspace*{-6mm}
\end{table*}

\begin{figure}
    \centering
    \includegraphics[width=0.48\textwidth]{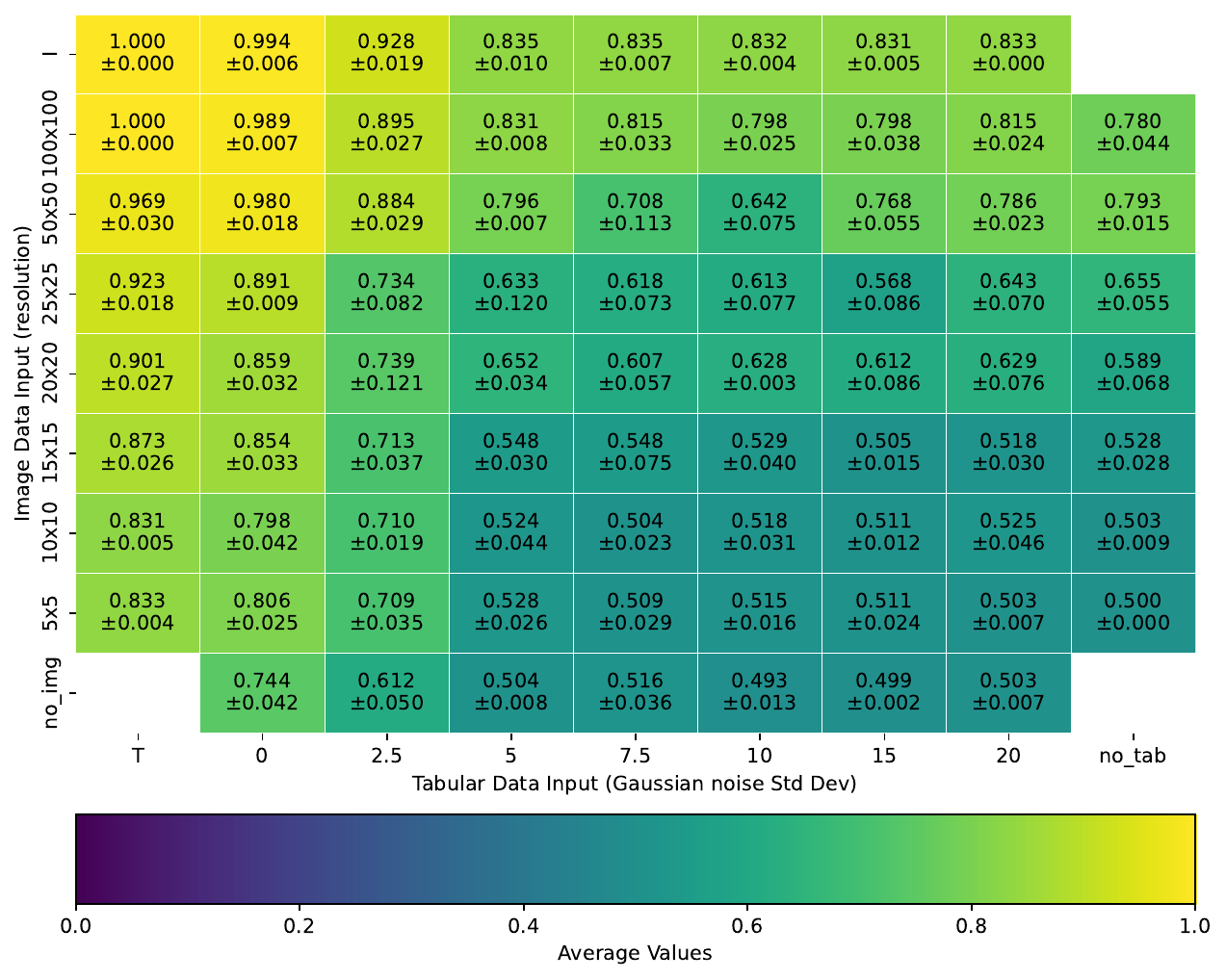}
    \vspace*{-6mm}
    \caption{Performance result matrix for the Multiclass Problem using Balanced Accuracy~(BAcc). Rows represent imaging inputs with different resolutions. Columns represent tabular inputs with different standard deviations used in the added Gaussian noise. The results consist of the average BAcc and its standard deviation over 5-fold cross validation.}
    \label{fig:fusion_results_multiclass}
\vspace*{-4mm}
\end{figure}

In Figure~\ref{fig:fusion_results_multiclass}, the obtained results for MultiFIX using end-to-end DL training are presented according to the combination of the available image and tabular inputs. I and T refer to the already extracted representative feature for image and tabular data, respectively. The last row and last show the predictive potential of each \emph{single} modality to predict the final label, for comparison purposes. Statistical tests showed a p-value lower than 0.05 between both image \textit{vs.} fusion and tabular \textit{vs.} fusion performance for the best-performing models.

\subsubsection{Interpretability Results}

\begin{figure}
    \centering
    \scalebox{0.95}{\includegraphics[width=0.5\textwidth]{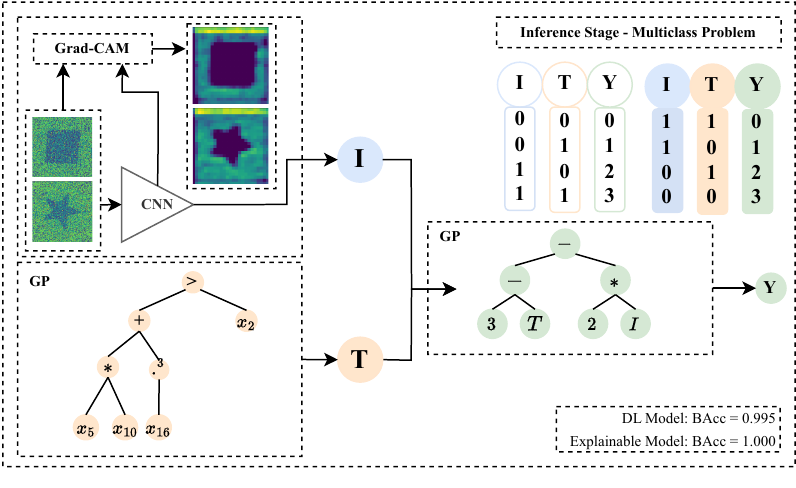}}
\vspace*{-8mm}
    \caption{Interpretability example for Multiclass Problem with image resolution $100\times100$ pixels and a tabular input with no Gaussian noise (standard deviation $0$). The new truth table for the Multiclass Problem is presented on the right, whilst the designed truth table is presented on the left.}
    \label{fig:multiclass_xai}
\vspace*{-7mm}
\end{figure}

An example of interpretability in the Inference Stage is demonstrated in Figure~\ref{fig:multiclass_xai} for an image input of resolution $100\times100$ pixels and a tabular input with no Gaussian noise (standard deviation $0$), for which the BAcc is 0.995 in the DL model, and 1.000 after replacing the tabular and fusion blocks with GP-GOMEA derived expressions. Input image samples and respective visual explanations are given to understand the main input contributions for the image extracted feature, I. Tabular feature T is now represented by the symbolic expression $((x_5*x_{10}) + x_{16}^3) > x_2$. The fusion of the features from each modality to predict the Multiclass target label is represented with the symbolic expression $3-T-2I$.

\subsubsection{Discussion}

A general decrease in predictive performance can be observed in Figure~\ref{fig:fusion_results_multiclass} when the noise level increases. Single modality performance demonstrated in the last row and last column of the performance heatmap, is, in general, lower than the fusion results. Thus, MultiFIX is able to outperform the single modality models as long as the model for the added modality has some predictive performance by itself.

Furthermore, the first row and first column of the performance heatmap show that the best performance average values are achieved when one of the representative features is given with no need for a feature engineering block.

Regarding interpretability, Figure~\ref{fig:multiclass_xai} showcases the use of the previously mentioned explainability methods. The symbolic expression that combines the induced features of the two modalities correctly predicts the target label. The table represented by the model, although different from the one used to construct the data present in Table~\ref{tab:multiclass_table}, still represents the same information. Since the relevant features were learned in an unsupervised fashion, these can be inverted compared to the constructed ones: in this example, the image feature $I$ represents the presence of a star in the picture and not the presence of a square.

\subsection{Multifeature Problem}
\label{subsec:and_or}
In the previous problem, the target label depends on a singular feature per modality. However, in real-world datasets, matters tend to be more complex, and, thus, more than one representative feature is often needed from each modality. We, therefore, consider an additional problem in which the target label is obtained from the Boolean operation shown in Equation~\ref{eq:multilabel}, using two features from the imaging modality and two features from the tabular modality.

\vspace*{-3mm}
\begin{equation}
    Y = AND(AND(ft_{circle},ft_A),OR(ft_{rectangle},ft_B))
    \label{eq:multilabel}
\end{equation}

\subsubsection{Data Description}
\begin{figure*}
    \centering
    \includegraphics[width=0.99\textwidth]{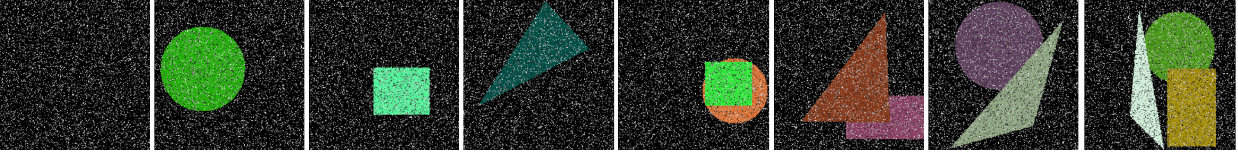}
    \vspace*{-4mm}
    \caption{Representative samples of input images for the Multifeature Problem.}
    \label{fig:multilabel_samples}
\vspace*{-3mm}
\end{figure*}

Image data was generated with three possible shapes: rectangles, circles, and triangles. Each shape appears in each image with 50\% chance. It is, therefore, possible to have 0, 1, 2, or 3 shapes in each image, with random sizes and colours. Random noise was added to all the images by randomly mutating the color of 10,000 randomly chosen pixels. Figure~\ref{fig:multilabel_samples} illustrates the created image data. Tabular data consisted of 10 continuous features uniformly randomly sampled between $0$ and $1$. The subsequently engineered features follow Equation~\ref{eq:tab_multilabel}. Since this problem requires multiple features to be engineered, 1000 samples were used.

\vspace*{-3mm}
\begin{equation}
    \begin{cases}
        A = 1, \text{ if } ft_0 + ft_1 + ft_2 > 1.5 \\
        B = 1, \text{ if } ft_3 + 2ft_4 + ft_5 > 2 \\
        C = 1, \text{ if } ft_6 + 3ft_7 + ft_8 > 2.5
    \end{cases}
    \label{eq:tab_multilabel}
\end{equation}

\subsubsection{Experimental Design}
For the image analysis, the pre-trained Resnet is now used in a multilabel fashion, being trained to predict the presence of rectangles, circles, and triangles in the images. For the tabular block, an MLP with a bottleneck of three nodes is used to detect the three possible features A, B, and C. 

HPO is used to find the optimal LR and WD with the grid available in Table~\ref{tab:multilabel_grid}. NAS was not used in this example due to the large amount of data samples required to train the models. However, results achieved through manual adjustment of the fusion block architecture were enough to find a benefit in using the two modalities simultaneously. The interpretable methods are applied to explain each feature individually. The fusion block is replaced with a symbolic expression generated by GP-GOMEA.

\begin{table}[]
\scalebox{0.81}{
\begin{tabular}{l|ll}
\multirow{2}{*}{\textbf{HPO}} & \textbf{LR} & {[}1e-2, 1e-3, 1e-4, 1e-5{]} \\ \cline{2-3} 
 & \textbf{WD} & {[}1e-2, 1e-3, 1e-4, 0{]} \\
\end{tabular}}
\caption{Optimisation Grid for Multifeature Problem and Melanoma Detection. The best set of parameters is chosen according to the loss (lowest average $\pm$ standard deviation calculated over the 5 folds).}
\label{tab:multilabel_grid}
\vspace*{-6mm}
\end{table}

\subsubsection{Performance Results}

In Table ~\ref{tab:results_multilabel} the performance results for the Multifeature Problem are presented, including the performance when only one of the modalities is used. The performance for supervised feature extraction is presented in Table~\ref{tab:ft_multilabel}. Statistical tests showed a p-value lower than 0.05 between both image \textit{vs.} fusion and tabular \textit{vs.} fusion performance.

\begin{table}
\scalebox{0.81}{
\begin{tabular}{l|ccc}
\textbf{} & \textbf{Image} & \textbf{Tabular} & \textbf{Fusion} \\ \hline
%\textbf{AUC} & 0.758 $\pm$ 0.031 & 0.813 $\pm$ 0.032 & 0.901 $\pm$ 0.047 \\
\textbf{BAcc} & 0.586 $\pm$ 0.055 & 0.590 $\pm$ 0.047 & 0.789 $\pm$ 0.095
\end{tabular}}
\caption{Performance results (average $\pm$ standard deviation calculated over the 5 folds) for the Multifeature Problem.}
\label{tab:results_multilabel}
\vspace*{-6mm}
\end{table}

\begin{table}
\scalebox{0.81}{
\begin{tabular}{l|cc}
    \textbf{} & \textbf{Image} & \textbf{Tabular}\\ \hline
    %\textbf{AUC} & 0.992 $\pm$ 0.004 & 0.996 $\pm$ 0.002 \\
    \textbf{BAcc}  & 0.971 $\pm$ 0.009 & 0.953 $\pm$ 0.015 
\end{tabular}}
\caption{Performance results (average $\pm$ std.dev. calculated over 5 folds) for the supervised feature extraction of each input modality separately in the Multifeature Problem.}
\label{tab:ft_multilabel}
\vspace*{-6mm}
\end{table}

\subsubsection{Interpretability Results}
\begin{figure}
    \centering
    \scalebox{0.95}{\includegraphics[width=0.5\textwidth]{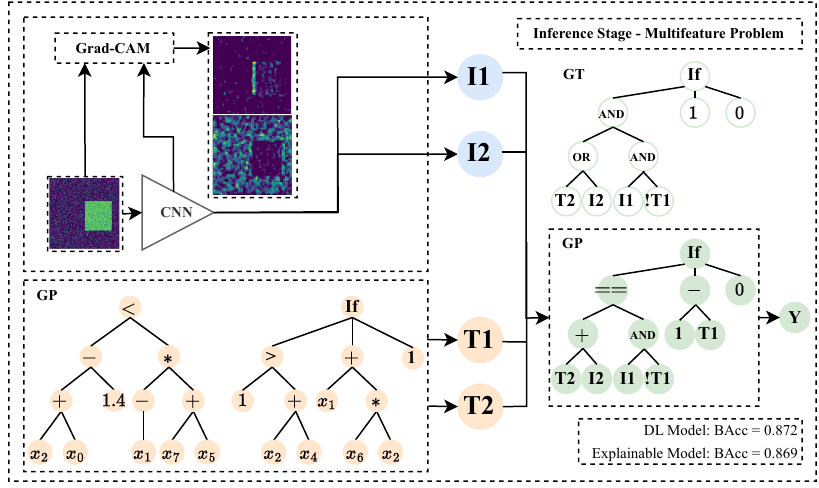}}
\vspace*{-7mm}
    \caption{Interpretability example for the Multifeature Problem and an image sample including a rectangle. The equivalent ground truth~(GT) symbolic expression is shown as a comparison to the actual expression selected by GP-GOMEA.}
    \label{fig:multilabel_xai}
\vspace*{-2mm}
\end{figure}

In figure~\ref{fig:multilabel_xai} the interpretable model is illustrated. Since this problem was structured to use two features from each data modality, Grad-CAM now presents two visual explanations per sample, one for each node after feature extraction. Similarly, GP-GOMEA is now used to evolve two symbolic expressions for the tabular feature engineering block. 

\subsubsection{Discussion}
With the end-to-end learning approach for MultiFIX, a significantly higher BAcc value could be obtained compared to using only single modalities as input. However, the supervised feature engineering performance of both modalities, presented in Table~\ref{tab:ft_multilabel}, would suggest that fusion would be able to achieve higher performance if the intermediate features were learned more accurately in the unsupervised setting as used in MultiFIX.

The interpretable model is showcased in Figure~\ref{fig:multilabel_xai}. The symbolic expression that represents the tabular feature T1 shows resemblances to the equation used to generate feature A, available in Equation~\ref{eq:tab_multilabel} since it is equivalent to $ft_0 + ft_1(x_7 + x_5) + x_2 < 1.4$. For the fusion block, a symbolic expression with a similar performance to the DL model could be obtained with GP-GOMEA. Furthermore, the obtained expression for the fusion shows high similarity to the ground truth expression: the relationship between I1 and the negation of T1 matches the $AND$ operation from the original expression, indicating that I1 is correlated with $ft_{circle}$ and T1 is correlated with $1-ft_A$; the sum of T1 and I2 outputs a similar value of the $OR$ operation between $ft_{rectangle}$ and $ft_B$, with the exception of the entry where both features are $1$, which constitutes one of the flaws in the model learned within MultiFIX using DL. Lastly, the comparison between the two expressions is made with the operation $==$ instead of the $AND$ operator, which outputs a similar set of values with the exception of both expressions being $0$, which demonstrates another flaw in the trained model. The explanations provided by the model allow us to understand in which cases the end-to-end learning approach failed, and, thus, to understand for which samples the developed model does not perform well. In other words, while the DL-based learning did not lead to satisfactory results, the explanation part of MultiFIX did.

\subsection{XOR Problem}
\label{subsec:xor}
In this problem, one feature from each modality is used to compute the binary operation $XOR(ft_{img},ft_{tab})$. This is considered to be extreme because each modality by itself does not have any correlation with the end-point, as demonstrated in Table~\ref{tab:xor_table}.

\begin{table}[]
    \centering
    \scalebox{0.81}{
    \begin{tabular}{c c|c}
         $ft_{circle}$ & $ft_{A}$ & $XOR$\\ \hline
         0 & 0 & 0 \\
         0 & 1 & 1 \\
         1 & 0 & 1 \\
         1 & 1 & 0
    \end{tabular}}
    \caption{End-point prediction for the XOR Problem.}
    \label{tab:xor_table}
\vspace*{-6mm}
\end{table}

\subsubsection{Data Description}
\begin{figure}
    \centering
    \scalebox{0.95}{\includegraphics[width=0.48\textwidth]{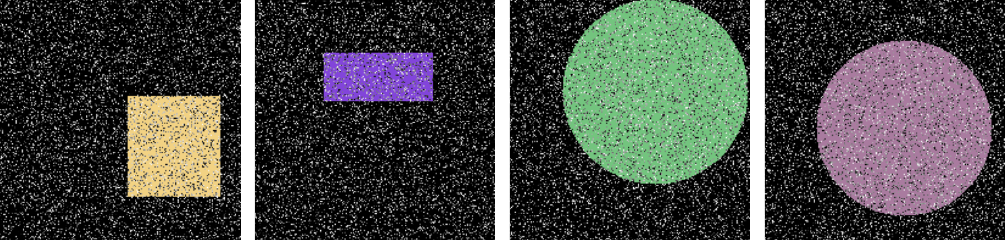}}
    \vspace*{-3mm}
    \caption{Representative samples of images used as input for the $XOR$ problem.}
    \label{fig:xor_samples}
    \vspace*{-2mm}
\end{figure}

The image data consists of samples with either a rectangle (label $0$) or a circle (label $1$). The shapes can vary in size and color, and noise was added to all images as in the Multifeature Problem. For an illustration, see Figure~\ref{fig:xor_samples}. The tabular data is built with 15 continuous features between $0$ and $1$. The representative tabular feature follows feature A of Equation~\ref{eq:tab_multilabel}. Both data modalities were generated automatically and the total number of samples is 1000.

\subsubsection{Experimental Design and Adaptations after Preliminary Results}

The same experimental design as used for the previous synthetic problems was first used. However, preliminary results showed that, with end-to-end learning of the three NN blocks, the features from each modality could not be properly extracted and combined to obtain the correct prediction. We found that the tabular block is not capable of learning, predicting a similar value for the extracted feature for all the samples. Consequently, overfitting occurs in the image block, adapting the output according to the fixed tabular feature to obtain the best possible prediction in the training stage and, thus, only using half of the truth table.

To solve such extreme problems, it is crucial to force the feature-inducing NN blocks for both modalities to learn simultaneously. We, therefore, propose an additional pre-training step for the image block: an Autoencoder~(AE) is trained in an unsupervised fashion to learn the most representative features of the image input in a linear latent space. The image feature engineering block of the multimodal pipeline is replaced with the pre-trained encoder architecture and an additional fully connected layer that can be trained in MultiFIX to select the representative feature. The encoder block is frozen for the first epochs of the training stage but afterward allowed to be trained as well, enabling further fine-tuning to support the final prediction task. This adds a new hyper-parameter to optimize: the epoch at which the encoder block is de-frozen in order to fully train the XOR problem.

This adaptation of the experimental design is evaluated and again compared to the single modality baselines. The image modality baseline also includes a version in which the encoder architecture is frozen, to assess the predictive potential of the features present in the latent representation of the AE.

HPO includes the de-freezing epoch for the encoder blocks, additionally to LR and WD. NAS was not used due to not only the large amount of data but also due to the computation expense of including the AE in the training pipeline. However, the results achieved through manual adjustment of the fusion block architecture were enough to draw conclusions about the benefit of using the two modalities simultaneously.

\begin{table}[]
\scalebox{0.81}{
\begin{tabular}{l|ll}
\multirow{3}{*}{\textbf{HPO}} & \textbf{LR} & {[}1e-2, 5e-3, 1e-3, 5e-4{]} \\ \cline{2-3} 
 & \textbf{WD} & {[}1e-2, 1e-3, 1e-4, 0{]} \\ \cline{2-3}
 & \textbf{De-freezing epoch} & {[}20, 30, 40, 50{]} \\
\end{tabular}}
\caption{Optimisation Grid for XOR Problem. The best set of parameters is chosen according to the loss (lowest average $\pm$ standard deviation calculated over the 5 folds)}
\label{tab:xor_grid}
\vspace*{-3mm}
\end{table}

\subsubsection{Performance Results}

Table ~\ref{tab:xor_results} shows the performance results for the XOR Problem, including the baseline performance of only using one of the modalities. Table~\ref{tab:ft_xor} shows the supervised feature extraction of each modality. Statistical tests showed a p-value lower than 0.05 between both image \textit{vs.} fusion and tabular \textit{vs.} fusion performance.

\begin{table}[]
\scalebox{0.81}{
\begin{tabular}{l|cc}
    \textbf{Input} & Image & Tabular\\ \hline
    %\textbf{AUC} & 0.999 $\pm$ 0.001 & 0.998 $\pm$ 0.000 \\
    \textbf{BAcc}  & 0.982 $\pm$ 0.007 & 0.975 $\pm$ 0.006
\end{tabular}}
\caption{Baseline performance results (average $\pm$ standard deviation calculated over the 5 folds) for the supervised feature extraction of each input modality in the XOR problem.}
\label{tab:ft_xor}
\vspace*{-3mm}
\end{table}

\begin{table*}[]
\scalebox{0.81}{
\begin{tabular}{l|cccccc}
    \textbf{Input} & Image & Tabular & Fusion & AE Fusion & Frozen AE Fusion & HPO AE Fusion\\ \hline
    %\textbf{AUC} & 0.503 $\pm$ 0.022 & 0.487 $\pm$ 0.042 & 0.998 $\pm$ 0.000 \\
    \textbf{BAcc}  & 0.509 $\pm$ 0.023 & 0.500 $\pm$ 0.012 & 0.495 $\pm$ 0.018 & 0.652 $\pm$ 0.190 & 0.778 $\pm$ 0.123 & \textbf{0.965 $\pm$ 0.014}
\end{tabular}}
\caption{Performance results (average $\pm$ st.dev. calculated over 5 folds) for the XOR problem using different training approaches. Image and Tabular correspond to using only one of the modalities; Fusion corresponds to using MultiFIX with direct end-to-end-learning; AE Fusion corresponds to the addition of a pre-trained encoder in the image feature engineering block; Frozen AE Fusion uses the encoder without trainable parameters; HPO AE Fusion corresponds to using HPO to adjust LR, WD, and the epoch from which the encoder block is trained along with the rest of the architecture.}
\label{tab:xor_results}
\vspace*{-6mm}
\end{table*}

\subsubsection{Interpretability Results}
\begin{figure}
    \centering
    \scalebox{0.95}{\includegraphics[width=0.5\textwidth]{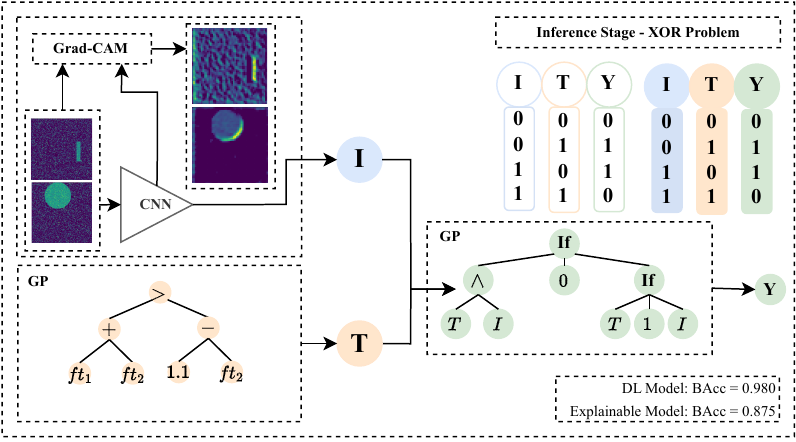}}
    \vspace*{-7mm}
    \caption{Interpretability example for the XOR Problem.}
    \label{fig:xor_xai}
    \vspace*{-5mm}
\end{figure}

The interpretable model is illustrated in Figure~\ref{fig:xor_xai}. Note that we did not allow GP-GOMEA to use XOR, otherwise the search would have been trivial. The truth table for the symbolic expression obtained with GP-GOMEA is equivalent to the truth table of the XOR. The visual explanations from the test samples show that the features that distinguish a circle from a rectangle are focused on the edge characteristics of each shape: a straight line is correlated with a rectangle, whilst a curved line indicates the presence of a circle. The tabular feature represented by the symbolic expression $2ft_2 + ft_1 > 1.1$ uses two of the features to compute feature A in Equation~\ref{eq:tab_multilabel}. 

\subsubsection{Discussion}

The results indicate the need for pre-training of the image feature engineering block using an AE. Without the AE, the resulting model has no predictive value when using both data modalities. The best performance was only achieved by freezing temporarily the encoder, hence forcing the tabular feature engineering block to start the learning process in combination with the preliminary image features learned with the AE, and in a second stage de-freeze the encoder block to fully train the model.

Regarding interpretability, GP-GOMEA can find an expression that captures the predicted features and computes the XOR of the latter. These results show that it is possible to simultaneously learn features from data modalities that, if used separately, cannot predict the final label. However, an end-to-end DL approach was insufficient as additional steps were required. This is likely due to the use of backpropagation to train the neural networks and the fact that the XOR operation makes it very difficult to propagate errors made in one modality to the right feature-inducing network block.

Contrary to the previous synthetic problems, there was a decrease in performance when replacing the tabular and fusion blocks with symbolic expressions generated with GP-GOMEA. Since the fusion expression mimics the intended target, one can justify this accuracy decrease with the limited representation of the engineered features. Since T and I are binary features, a threshold of 0.50 is assumed to binarise them. Thus, there is an inherent possible bias from the DL pipeline that uses probability values for the extracted features. Optimising the threshold could provide a solution.
%complex and computationally expensive methods were required to force both modalities to train and avoid local minima in the loss landscape.

\subsection{Three-Gated XOR Problem}
\label{subsec:3_xor}
One last example was studied that increases complexity even further. Several features from each feature engineering block are needed to predict the target label, while individually having no predictive value. Specifically, an XOR with three possible inputs was used, using two features from the image data - the presence of a circle, $ft_{circle}$, and the presence of a triangle, $ft_{triangle}$, and one tabular feature - feature A, $ft_A$, computed as described in Equation~\ref{eq:tab_multilabel}. Subsection~\ref{subsec:and_or} presents a detailed description of the input data.

\subsubsection{Experimental Design}
As this problem is even harder, the modifications described in the two-gated XOR were used for this problem also by pre-training the image feature engineering block with an AE and temporarily freezing it in the multimodal pipeline.

%\textcolor{red}{explain in depth why it's not working}

Even so, we were not able to force the model to simultaneously learn the three engineered features from the available modalities and compute the three-gated XOR. As mentioned earlier for the XOR problem, we believe that the developed architecture, specifically the bottleneck of each engineering block to extract the representative features, may cause issues with the use of backpropagation as it is forced to assign errors due to one modality to the right feature-inducing block, which is not what backpropagation does. Furthermore, input samples with the same representative features on one modality can belong to either one of the classes, which forces the two modalities to learn simultaneously to make the right prediction. This may expose the model to opposite gradients for examples that look exactly the same in the feature engineering branches. This problem was tackled in the XOR Problem, Section~\ref{subsec:xor}, by pre-training the image block with an AE. However, since the three-gated XOR increases the chance of stagnation in the training process, which can be related to saddle points in the loss landscape, pre-training the image feature engineering block is not sufficient to solve the problem. One possibility to address this issue is the use of more sophisticated optimisation of the NNs without using backpropagation, which could well be an excellent opportunity for methods from the evolutionary computation community.

This problem identifies a clear limitation of MultiFIX when it comes to using several features from different modalities that by themselves do not have any predictive value towards the target label and using modern NN approaches. However, the extent to which this is a real-world hurdle is unclear as one would expect that typically the different data sources used as input are, \textit{a priori}, correlated to the target to some extent. For this reason, we finally turn to a real-world example.

\section{Application - Melanoma Dataset}
Data was sampled from the publicly available ISIC 2020 Challenge Dataset~\cite{melanoma_dataset}. A subset of 1000 samples was used to detect melanoma, containing dermoscopic images with skin lesions and clinical tabular data with the following features: age, anatomical location of the lesion, and sex. Cases with a malignant diagnosis label, were histopathologically verified. Benign-labeled cases went through an expert agreement, longitudinal follow-up, or histopathology. Image data was cropped to be squared and resampled to a standard resolution of $200\times200$ pixels. Regarding the tabular features, the anatomical location of the lesion was one-hot encoded.

\subsection{Experimental Design}
In a real-world application the key features needed from each modality are not known \textit{a priori}. Thus, multiple possibilities for the number of nodes in the bottleneck of each feature engineering block were considered with the premise that the best-performing architecture would have learned the optimal number of representative features for each modality. As a baseline, single-modality performance was evaluated. HPO is used to choose the optimal LR and WD, as in the previous approaches, using the grid from Table~\ref{tab:multilabel_grid}.

\subsection{Performance Results}

Table~\ref{tab:results_melanoma} shows the performance results in terms of BAcc. These results show similar performance values for most of the developed models, except for the model that predicts the target label using the tabular data exclusively. Several combinations of output nodes for the feature engineering blocks are shown. Statistical tests showed a p-value lower than 0.05 between tabular \textit{vs.} fusion performance.

\begin{table}[]
\begin{center}   
\scalebox{0.81}{\begin{tabular}{l|c|c}
 & \textbf{Image} & \textbf{Tabular} \\\hline
\textbf{BAcc} & 0.806 $\pm$ 0.027 & 0.627 $\pm$ 0.039\\
\end{tabular}}
\\\ \\
\tabcolsep=1mm
\scalebox{0.81}{\begin{tabular}{l|ccccc}
\textbf{Fusion} &  \multicolumn{1}{c|}{\textbf{1I + 1T}} & \multicolumn{1}{c|}{\textbf{2I + 2T}} & \multicolumn{1}{c|}{\textbf{3I + 3T}} & \multicolumn{1}{c|}{\textbf{4I + 4T}} & \textbf{5I + 5T} \\ \hline
%\textbf{AUC} & \multicolumn{1}{c|}{0.894 $\pm$ 0.037} & \multicolumn{1}{c|}{0.670 $\pm$ 0.052} & \multicolumn{1}{c|}{0.868 $\pm$ 0.045} & \multicolumn{1}{c|}{0.877 $\pm$ 0.039} & \multicolumn{1}{c|}{0.871 $\pm$ 0.035} & \multicolumn{1}{c|}{0.881 $\pm$ 0.044} & \multicolumn{1}{c|}{0.882 $\pm$ 0.043} \\
\textbf{BAcc} & \multicolumn{1}{c|}{0.789 $\pm$ 0.048} & \multicolumn{1}{c|}{0.812 $\pm$ 0.040} & \multicolumn{1}{c|}{0.797 $\pm$ 0.034} & \multicolumn{1}{c|}{0.787 $\pm$ 0.040} & \multicolumn{1}{c}{0.795 $\pm$ 0.042}
\end{tabular}}
\end{center}
\caption{Performance results (BAcc average $\pm$ std.dev. calculated over 5 folds) for Melanoma Detection with a different number of representative features for each data type.}
\label{tab:results_melanoma}
\vspace*{-7mm}
\end{table}

\subsection{Interpretability Results}
\begin{figure}
    \centering
    \scalebox{0.95}{\includegraphics[width=0.5\textwidth]{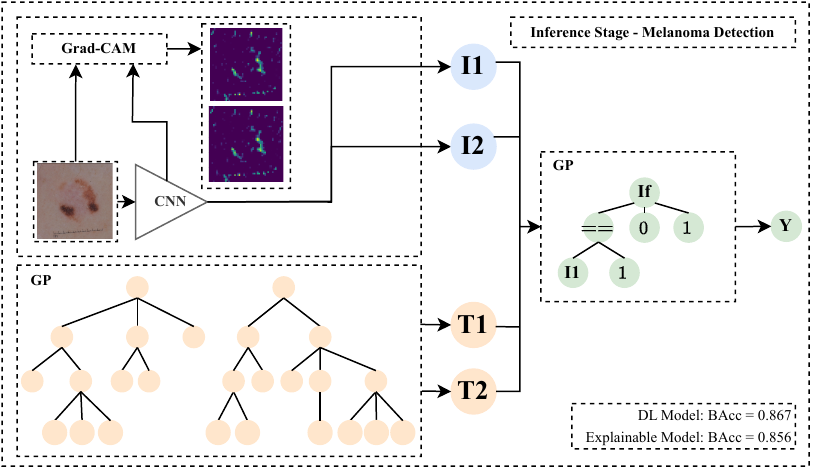}}
    \vspace*{-7mm}
    \caption{Interpretability example for the Melanoma Dataset.}
    \label{fig:xai_melanoma}
    \vspace*{-5mm}
\end{figure}

An example of the interpretability stage is demonstrated in Figure~\ref{fig:xai_melanoma}. The model chosen to demonstrate is the best-performing model, with two engineered features from each data modality. 

\subsection{Discussion}

Unfortunately, the combination of the image and tabular modalities does not contribute to a better prediction than when only image features are used. The addition of a modality with a predictive performance of $0.627\pm0.039$ did not deteriorate predictive ability as the tabular features were disregarded in the final model.

Regarding the interpretability of the developed pipeline, the GradCAM explanations for the extracted image features are very similar. The feature values are also very similar, which indicates that the extracted features may be redundant. The final GP tree for the fusion block shows that this is indeed the case, as essentially the final prediction is equivalent to engineered image feature I1. Tabular engineering features used clinical information, mainly the age of diagnosis. However, since the tabular features do not contribute to the prediction of melanoma, further analysis was not necessary.

Altogether, unfortunately, this multimodal dataset turned out to have  little synergistic value of the different datatypes for the final prediction. However, a key part of the reason why we presented MultiFIX is the ability to explain learned models. In this case, we could easily identify that the tabular features have no added value and that only one image feature is sufficient to get the best performance. In other words, similar to the Multifeature Problem where the NN-based training was not optimal, the decomposition of explanations along the features induced from different data types and the final prediction, was insightful.

\vspace*{-2mm}
\section{Conclusions and Future Work}
In this paper, we have presented MultiFIX, the first XAI-friendly feature inducing approach to building predictive models from multimodal data. The architecture of MultiFIX consists of a feature-construction model for each type of data, and one more model that combines (i.e., fuses) these features to make the final prediction. For ease of learning, and as a baseline, we have studied the use of NN architectures for each of the feature-construction block as well as an NN architecture for the fusion block, so that the entire architecture can be learned end-to-end using typical backpropagation techniques from modern deep learning. Afterward, parts can be explained using post-hoc explanation techniques, or, whenever appropriate, be replaced entirely be inherently interpretable models. As trust, explainability, and accountability become ever more important with the advancement of AI, especially in domains such as the medical domain, approaches like MultiFIX are expected to play a crucial role in future employment of AI models in practice.

While we have observed on several synthetic datasets and one real-world dataset that MultiFIX can indeed be used to create piecewise explainable models that have high prediction accuracy, the use of NN training techniques such as backpropagation and pre-trained autoencoders to first train multiple NNs and subsequently explain them or replace them with inherently interpretable models, were found to have their limitations. In particular, as the joint dependence between modalities to make highly accurate predictions increases, the proposed architecture cannot be trained well with standard NN training techniques anymore.

Evolutionary methods such as GP already play an important part in MultiFIX, to provide inherently explainable models. Still, the observed limitations of standard NN training techniques call for an even deeper integration of powerful optimization techniques. These may include modern EAs that do not rely on (backpropagated) gradients to scale MultiFIX to effectively build more complex models with potential strong interactions between features of different datatypes. We believe this to be a unique challenge and opportunity for evolutionary machine learning research toward the future.

\vspace*{-2mm}
\begin{acks}
The authors would like to thank Krzysztof Krawiec for insightful discussions regarding Section~\ref{subsec:3_xor}.
This research is part of the "Uitlegbare Kunstmatige Intelligentie" project funded by the Stichting Gieskes-Strijbis Fonds.
\end{acks}

%%
%% The next two lines define the bibliography style to be used, and
%% the bibliography file.
\bibliographystyle{ACM-Reference-Format}
\bibliography{sample-base}

%%
%% If your work has an appendix, this is the place to put it.
%\appendix

\end{document}